# Unsupervised Pre-Training for 3D Leaf Instance Segmentation

Gianmarco Roggiolani    Federico Magistri    Tiziano Guadagnino    Jens Behley    Cyrill Stachniss

*Abstract*—Crops for food, feed, fiber, and fuel are key natural resources for our society. Monitoring plants and measuring their traits is an important task in agriculture often referred to as plant phenotyping. Traditionally, this task is done manually, which is time- and labor-intensive. Robots can automate phenotyping providing reproducible and high-frequency measurements. Today's perception systems use deep learning to interpret these measurements, but require a substantial amount of annotated data to work well. Obtaining such labels is challenging as it often requires background knowledge on the side of the labelers. This paper addresses the problem of reducing the labeling effort required to perform leaf instance segmentation on 3D point clouds, which is a first step toward phenotyping in 3D. Separating all leaves allows us to count them and compute relevant traits as their areas, lengths, and widths. We propose a novel self-supervised task-specific pre-training approach to initialize the backbone of a network for leaf instance segmentation. We also introduce a novel automatic postprocessing that considers the difficulty of correctly segmenting the points close to the stem, where all the leaves petiole overlap. The experiments presented in this paper suggest that our approach boosts the performance over all the investigated scenarios. We also evaluate the embeddings to assess the quality of the fully unsupervised approach and see a higher performance of our domain-specific postprocessing.

## I. Introduction

OUR society heavily relies on crop production for providing food, feed, fiber, and fuel. The demand for biomass is constantly increasing and is expected to grow even further over the next decades. This causes serious challenges to the agricultural systems to meet such demands and simultaneously produce biomass in a sustainable manner.

Agricultural robots have the potential to change crop production by automating processes, enable fine-grained and high-resolution monitoring, and executing targeted intervention tasks. In recent years, many studies have been published analyzing applications in the context of agricultural robotics, intending to increase yield and reduce human labor and agro-chemical inputs [1]. Robotic systems can realize sustainable weeding [2], control pesticide usage [3], detect fruits for yield prediction [4], realize selected harvesting [5], and perform phenotyping [6][7].

Phenotyping refers to measuring the observable traits of an organism. It plays a crucial role in crop production and plant breeding, where monitoring the appearance and performance of different varieties is key; today, this task still involves lots of human labor. This poses a limit on the number of measures and the time required to analyze them. Estimating the number of leaves of individual plants and their size is a typical task in this context. In particular, the number of leaves and their area can provide information about the yield [8], health [9], and need for nutrients or pesticides [10].

The phenotyping task has been addressed in the 3D scenario using supervised networks on point clouds from 3D LiDAR sensors or multi-view images. Most of these approaches build on general-purpose instance segmentation approaches [11][12]. Thus, the initial struggle to gather the plant traits is still present in order to build a training dataset. To fully automate the phenotyping process, the robot needs a robust perception system able to acquire phenotypic traits in an automated and repeatable fashion.

In this paper, we aim to reduce the amount of labeled data needed to achieve state-of-the-art performance on leaf instance segmentation via a novel self-supervised pre-training approach. Supervised pre-training on point clouds [14] is still far behind compared to the image-based scenario, where it is common practice to pre-train networks on general-purpose datasets like ImageNet [15] or MS COCO [16]. Unsupervised 3D pre-training is less common and usually application-specific [17], we also make our pre-training domain-specific and also task-specific.

The main contribution of this paper is a 3D self-supervised pre-training to differentiate each leaf of each plant. We design domain-specific augmentations and exploit task and domain knowledge to build a more specific self-supervised loss. We fine-tune on labeled data to show the improvement achieved thanks to our pre-training. We also propose novel automatic postprocessing of the self-supervised output, taking into consideration the difficulties of differentiating individual leaves — especially in the stem region — and reducing the impact on the final performance. In sum, we show that: (i) our task-specific pre-training improves the performance on leaf instance segmentation and reduces the amount of labeling required; (ii) distance information and number of points are more important than the use of a second view as in common contrastive learning; (iii) increasing the embedding size boosts the performance of our pre-training more than the performance of

Manuscript received: May 19, 2021; Revised: Jul 16, 2021; Accepted: Sep 15, 2021. This paper was recommended for publication by Editor Hyungpil Moon upon evaluation of the Associate Editor and Reviewers' comments.

All authors are with the University of Bonn, Germany. Cyrill Stachniss is additionally with the Department of Engineering Science at the University of Oxford, UK, and with the Lamarr Institute for Machine Learning and Artificial Intelligence, Germany.

This work has partially been funded by the Deutsche Forschungsgemeinschaft (DFG, German Research Foundation) under Germany's Excellence Strategy, EXC-2070 – 390732324 – PhenoRob.

Digital Object Identifier (DOI): see top of this page.

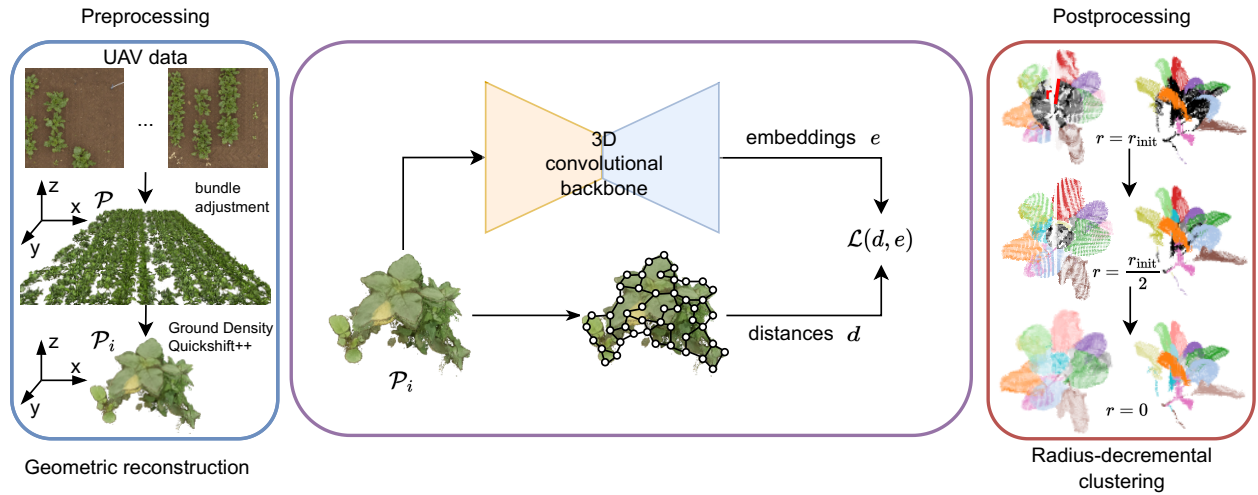

Fig. 1: Overview of our pipeline. In the preprocessing (left), we build the point cloud $\mathcal{P}$ from UAVs images using bundle adjustment and segment $\mathcal{P}$ using the method by Nelson et al. [13] into single clouds $\mathcal{P}_i$. These are the inputs of our network, which learn representations computing the loss $\mathcal{L}$ on per-point embeddings **e**. In the postprocessing (right) we exploit domain-specific knowledge to cluster the embeddings and distinguish each leaf, starting from outer points and progressively assigning points closer to the center.

the randomly initialized network; and (iv) our novel automatic domain-specific postprocessing achieves better performance with respect to common state-of-the-art methods.

## II. RELATED WORK

The problem of instance segmentation of leaves has often been addressed in the 2D domain, i.e., on RGB images, mainly using neural networks. Different works proposed convolutional neural networks (CNN) based on classical instance-segmentation architectures such as Mask-RCNN [18] and ERFNet [7]. In the 3D domain, there are two main paradigms: geometric approaches [19][20], which try to exploit the knowledge about the plants' structure, and learning-based approaches [12][21].

As for the first paradigm, Miao et al. [22] propose to extract the skeleton of the plant, run a first coarse segmentation, and then perform a fine segmentation based on morphological features. Jin et al. [23] follow a similar approach, using a growing algorithm to segment the stem after removing the ground points. The main disadvantage of purely geometric pipelines is that they cannot infer what is occluded and usually fail to distinguish close leaves.

Ao et al. [24] propose a two-stage approach based on PointCNN [25] to extract stem points used to fit 3D cylinders. Single leaves are extracted through a density-based algorithm [26] and postprocessed according to morphological characteristics. The approach of Ao et al. is in between the two main paradigms, since they rely on a deep learning approach to segment the stem, reducing the assumptions about the plant species and growing orientation. This means that they do not require fine annotations for the leaf instance to train the network, but also that they rely solely on the points' positions for clustering the leaves, making it hard to separate the leaves when occlusions or overlaps occur.

Han et al. [27] instead propose to use a deep learning approach to solve the whole problem, directly predicting instances whose instance can be computed via the mean-shift clustering algorithm [28]. Han et al. indicate as crucial for their success both, the filtering step in their preprocessing and the double supervision of the network, which aims to achieve segmentation and completion of the point cloud simultaneously. Compared to the previous approaches, end-to-end training like this one requires a lot of labeling effort.

A common way to reduce this effort is the pre-training of the network in a self-supervised fashion, such that it can converge faster and with less labeled data. Lately, self-supervised learning has focused on contrastive approaches [29][30], which augment the input to produce different views with the goal of producing features that are close for views coming from the same sample. Xie et al. [31] adapt the contrastive learning paradigm used for images [29][30] to the point cloud domain. After generating the two views, they find one positive example for each point and propose a new contrastive loss based on the InfoNCE loss [32]. Zhang et al. [33] also use a contrastive approach but rely on a momentum encoder to use more negative examples. They transform the two views into different input formats – points or voxels – and output a feature vector for each one to use in the contrastive loss. In contrast to these approaches, Wang et al. [34] provide a pre-text task to the network. They simulate occlusions erasing subsets of points and require the network to reconstruct the missing section. Alliegro et al. [35] also use a pre-text self-supervised task, a 3D puzzle. This auxiliary task makes the model more precise and robust also for out-of-domain generalization. The same results are obtained by Achituve et al. [36] with their deformation reconstruction, in which they dislocate points of the input and the network needs to predict the original locations.

Our work belongs to the class of self-supervised learning-based approaches, we define our loss in the feature space, and do not use a pre-text task. As all of the self-supervised methods, we aim to reduce the labeling effort by exploiting unlabeled data and our knowledge about the plant structure to better initialize a network for the leaf instance segmentation task. We address the task in 3D to make it more suitable to extract leaves' length, width, and shape, that are essential for phenotyping [37] and are harder to extract from images due to the leaves curl and curvature. The main contributions of this work are a spatially-informed unsupervised approach for pre-training and a novel automatic postprocessing.

## III. OUR APPROACH

We propose a new unsupervised approach to pre-train a deep neural network for leaf instance segmentation in 3D point clouds. The network is part of the pipeline shown in Fig. 1. The preprocessing computes a point cloud from UAVs images of the field and then extracts single plants leveraging the geometric approach explained in Sec. III-A. We do not only make our pre-training domain- and task-specific, we also apply agriculture-specific augmentations to the single point clouds before feeding them into the backbone. In the following sections, we refer to the augmented point clouds as views and show that our approach works both with one or two views. The backbone takes as input a sparse tensor $I \in \mathbb{R}^{N \times 6}$, where $N$ is the number of points in the point cloud and 6 is the features' dimension, 3 for the point position and 3 for the color. The backbone outputs per-point embeddings, which we can use to compute the unsupervised loss $\mathcal{L}$, to perform a fully unsupervised bottom-up leaf instance segmentation or as features to be refined by fine-tuning using labeled data. In the latter case, we load the pre-trained weights to initialize the backbone. It is usually followed by other layers to compute the final predictions used to obtain the instances.

### A. Preprocessing

We use RGB images of sugar beets collected by a UAV plus GPS information to build a dense point cloud $\mathcal{P}$ of the field via bundle adjustment [38]. The approach estimates 3D points locations and camera orientations that minimize the total reprojection error, then combines them into a single point cloud, with color information by projecting the color information from the corrisponding image pixel.

To separate the point cloud $\mathcal{P}$ into individual plants $\mathcal{P}_i$, we use Ground Density Quickshift++ by Nelson et al. [13], preserving the color. The algorithm first uses Quickshift++ [39] to initialize the clusters looking only at the x and y coordinates, and then refines these results considering the z-component. This two-step approach ensures that we do not separate stem points because of their different heights, and that leaves in the same area of the xy-plane are separated. We refer to the original paper [13] for more details.

### B. Augmentations

Our pipeline augments the individual input point clouds $\mathcal{P}_i$ via different transformations, which is crucial in unsupervised training because it helps the network to focus on relevant features. We use 3D versions of common 2D augmentations – rotation, translation, adding noise, and erasing points – and the below explained domain-specific augmentations to simulate leaf occlusion and distortion. These augmentations are applied during the unsupervised pre-training to obtain better weights for the network. If we use these weights to perform a fully unsupervised leaf instance segmentation, we need to discard them at inference time or preserve the information about the rotation applied, in order to correctly use our post-processing as described in Sec. III-D.

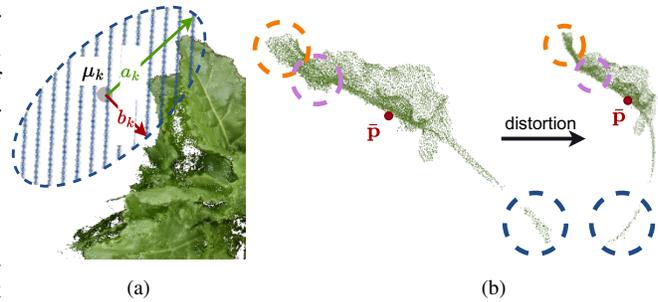

Fig. 2: Results of the augmentation proposed in Sec. III-B. Figure (a) generates leaf occlusion as all points in the generated ellipse with center $\mu_k$, and axes $a_k, b_k$ are removed. Figure (b) generates a leaf distortion and the dotted circles highlight corresponding areas at different distances from $\bar{\mathbf{p}}$.

**1) Leaf Occlusion:** this augmentation aims to cut out of the point cloud all the points falling into K 2D ellipses

$$\epsilon_k(x,y) = \frac{(x - \mu_{k,x})^2}{a_k^2} + \frac{(y - \mu_{k,y})^2}{b_k^2} - 1, \quad (1)$$

where $a_k$ and $b_k$ are the two axes of the ellipse $k$, and $\boldsymbol{\mu}_k = (\mu_{k,x}, \mu_{k,y})^\top$ is its center. This simulates the occlusion caused by the leaves of adjacent plants. We consider it a domain/specific variant of CutOut [40]. In our work, we use $K = 2$, since the plants are grown in rows and thus the occlusions are usually caused by plants in the same row. We randomly select either $a$ or $b$ to be the major axis, with the same probability to be selected. Then, we compute $\boldsymbol{\mu}_k$ as the furthest point in direction of the selected major axis with respect to the center of the plant $\bar{\mathbf{p}}_i = \frac{1}{|\mathcal{P}_i|} \sum_{\mathbf{q} \in \mathcal{P}_i} \mathbf{q}$.

We sample the dimensions of $a_k, b_k \sim \mathcal{U}\{0, \boldsymbol{\delta}\}$ from a uniform distribution $\mathcal{U}$, where $\boldsymbol{\delta} \in \mathbb{R}^2$ is a parameter defining the maximum length and width of the leaves according to the growth stage of the plant. We project the point cloud onto the xy-plane and remove all the points falling inside the ellipses. In the case of non-row-crops, we can change the parameters to have randomly spaced centers. The result of this augmentation is shown in Fig. 2 (a).

**2) Leaf Distortion:** this augmentation rotates the points to imitate the movement of the leaves caused by the wind. Instead of the classical rotation of the entire point cloud, we rotate each point according to its distance to the estimated plant center $\bar{\mathbf{p}}$.

Given maximum rotations $\boldsymbol{\theta}_{\max} = (\alpha_{\max}, \beta_{\max}, \gamma_{\max})^\top$ about the x-, y- and z-axes, we randomly sample their values $\boldsymbol{\theta} = (\alpha, \beta, \gamma)^\top$, i.e., $\alpha, \beta, \gamma \sim \mathcal{U}\{0, 1\}$. For each point $\mathbf{p} \in \mathcal{P}_i$, we compute $d_\mathbf{p} = ||\mathbf{p} - \bar{\mathbf{p}}_i||_2$ and define the per-point Euler angles $\boldsymbol{\theta}_\mathbf{p} = (\alpha_\mathbf{p}, \beta_\mathbf{p}, \gamma_\mathbf{p})^\top$ for the rotation as

$$\boldsymbol{\theta}_\mathbf{p} = \frac{d_\mathbf{p}}{\max_{\mathbf{q} \in \mathcal{P}_i} d_\mathbf{q}} \boldsymbol{\theta} \times \boldsymbol{\theta}_{\max}^\top. \quad (2)$$

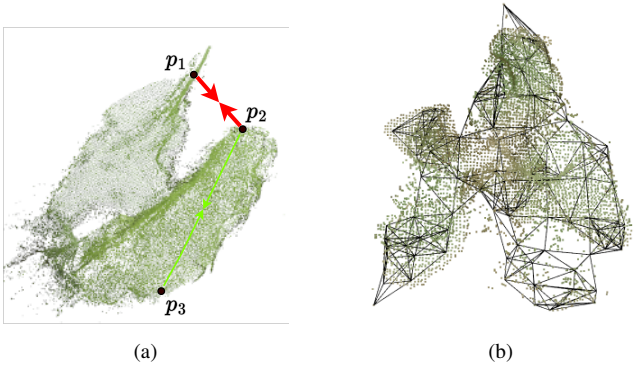

Fig. 3: In (a) $\mathbf{p_1}$ belongs to one leaf, while $\mathbf{p_2}$ and $\mathbf{p_3}$ belong to another. If we only use the euclidean distances, $\mathbf{p_2}$ will have an embedding more similar to $\mathbf{p_1}$ than to $\mathbf{p_3}$. In (b) we show the graph built over the down-sampled point cloud using $k = 7$ nearest neighbors and $\tau = 2\,\text{cm}$.

We can build for each point its rotation matrix

$$\mathbf{R_p}(\alpha_\mathbf{p}, \beta_\mathbf{p}, \gamma_\mathbf{p}) = \mathbf{R}_z(\gamma_\mathbf{p})\mathbf{R}_y(\beta_\mathbf{p})\mathbf{R}_x(\alpha_\mathbf{p}), \quad (3)$$

where $\mathbf{R}_a(\theta)$ is the rotation matrix around axis $a$ with angle $\theta$. We apply the distortion as $\mathbf{R_p}\,\mathbf{p}$, which leads to a distance-dependent rotation of points. Results of this augmentation are shown in Fig. 2 (b) and (c).

### C. Unsupervised Loss

We aim to learn per-point embeddings $\mathbf{e}_j^v \in \mathbb{R}^D$, with $j \in \{0, ..., N\}$, where $N$ is the number of points and $v$ is the view, to assign each point to one leaf. In particular, we want that the embeddings of the same point in the different views are identical.

In standard contrastive learning [29][30], we normalize the embeddings along the feature dimension as $\widehat{\mathbf{e}}_j = \frac{\mathbf{e}_j^v}{||\mathbf{e}_j^v||_2}$ and compute the cross-correlation between each pair of points. The loss can be expressed as

$$\mathcal{L} = \sum_{i,j=0}^{N} \mathbf{I}_{i,j} - \widehat{\mathbf{e}}_i^0 \widehat{\mathbf{e}}_j^{1\top}, \quad (4)$$

where $\mathbf{I} \in \mathbb{R}^{N \times N}$ is the identity matrix.

Since we aim to perform a bottom-up instance segmentation, we want to include spatial information and push the embeddings of close points to be similar.

To do so, we need to remove the identity matrix and use a matrix that carries the information about how similar each cosine similarity should be. This new loss function will be

$$\mathcal{L} = \sum_{i,j=0}^{N} \mathbf{S}_{i,j} - \widehat{\mathbf{e}}_i^0 \widehat{\mathbf{e}}_j^{1\top}, \quad (5)$$

where $\mathbf{S}_{i,j} \in \mathbb{R}^{N \times N}$ specifies how similar each pair of points should be given their distance. This means the loss can be used also with just one view. To include the spatial information we investigate Euclidean and graph distances. Since we augment the point clouds with translations and rotations, distances must be computed on the same view to be comparable.

For the Euclidean distance between points $\mathbf{p}_r, \mathbf{p}_c \in \mathcal{P}_i$, we simply set $\mathbf{D}_{r,c} = ||\mathbf{p}_r - \mathbf{p}_c||_2$. This makes the embeddings of points close in space similar, but when leaves overlap it can happen that points of different leaves are led to have more similar embeddings than points from the same leaf, as shown in Fig. 3(a).

To overcome this limitation, we propose to compute the graph distance on the plant. To do so, we need to create a graph $G = (V, E)$, where $V$ are the points and $E$ the edges given by the k-nearest neighbor graph—i.e. each point has edges to the k closest points if their distances are smaller than a threshold $\tau$. We build the graph with $k = 7$ and a maximum distance of $\tau = 2\,\text{cm}$ between connected points. The result of this operation is shown in Fig. 3(b). We then initialize a distance matrix $\widetilde{\mathbf{D}}$ as

$$\widetilde{\mathbf{D}}_{r,c} = \widetilde{\mathbf{D}}_{c,r} = \begin{cases} ||\mathbf{p}_r - \mathbf{p}_c||_2 & , \text{if } (r,c) \in E \\ \infty & , \text{otherwise}. \end{cases} \quad (6)$$

Afterwards, we use the Floyd-Warshall [41] algorithm to traverse the graph and fill the whole matrix. Each element of row $r$ and column $c$ in $\mathbf{D}$ is computed from $\widetilde{\mathbf{D}}$ as

$$\mathbf{D}_{r,c} = \mathbf{D}_{c,r} = \min_k(\widetilde{\mathbf{D}}_{r,k} + \widetilde{\mathbf{D}}_{k,c}). \quad (7)$$

In the end, $\mathbf{D}$ has a null diagonal, positive values if two nodes are connected, and $\infty$ otherwise. After filling up the matrix $\mathbf{D}$, we compute the similarities as

$$\mathbf{S}_{r,c} = \frac{1}{\mathbf{D}_{r,c} + \epsilon}, \quad (8)$$

where $\epsilon$ is an arbitrarily small value used to avoid numerical instability. We normalize the similarities as $\mathbf{S}_{r,c} = \frac{\mathbf{S}_{r,c}}{\max(\mathbf{S})}$. Thus, all points with zero distance will have a similarity of 1, and all other values will have a similarity inversely proportional to the distance between the points.

### D. Postprocessing

Instance segmentation tasks can be solved by predicting embeddings to cluster, or centers for each instance and offsets to their center for each point in the instance. Since it is hard to supervise the predictions of centers in the absence of labels, our pre-training produces embeddings. Thus, to evaluate the fully unsupervised approach, we need an embedding-based clustering postprocessing.

In the agricultural setting, points close to the center of the plant are more complex to assign correctly since many different leaves connect to the same stem. Considering this problem, we propose a novel automatic postprocessing that starts clustering the leaves from the outer points. When not specified, we cluster via agglomerative clustering [42], and use the cosine similarity to compute the similarity referred to as `sim`. Our postprocessing consists of two steps:

**1) Define radiuses to cut.** We compute the center of the plant $\bar{\mathbf{p}}$ and the distances to the farthests points on the x and y axes as

$$d_x = \max_{\mathbf{q} \in \mathcal{P}_i} |\mathbf{q}_x - \bar{\mathbf{p}}_x| \quad \text{and} \quad d_y = \max_{\mathbf{q} \in \mathcal{P}_i} |\mathbf{q}_y - \bar{\mathbf{p}}_y|. \quad (9)$$

We then define the initial radius of the area to cut as

$$r_{\text{init}} = \frac{\min(d_x, d_y)}{2}, \quad (10)$$

which allows us to distinguish the tips of the leaves as a starting point.

*2) Radius-decremental clustering.* At each step we decrease the size of the radius as

$$r = r_{\text{init}} - \frac{r_{\text{init}}}{S}, \quad (11)$$

where S is the number of steps we want to make in the postprocessing operations. We cluster only the embeddings of non-clustered points with distance from $\bar{\mathbf{p}}$ greater than $r$. For each of the new M clusters $\hat{\mathcal{C}}_k$ with $k \in \{1, M\}$, we compute the maximum similarity with respect to the existing clusters $\mathcal{C}$ using the embeddings mean, i.e.,

$$\hat{s} = \max_i \ \text{sim}\left(\frac{1}{|\hat{\mathcal{C}}_k|}\sum_{m \in \hat{\mathcal{C}}_k} \mathbf{e}_m, \frac{1}{|\mathcal{C}_i|}\sum_{j \in \mathcal{C}_i} \mathbf{e}_j\right). \quad (12)$$

If $\hat{s}$ is higher than a threshold $\gamma$, we merge the clusters, otherwise, we create a new cluster. We iterate this step until $r = 0$ and all points are assigned.

## IV. EXPERIMENTAL EVALUATION

The main focus of this work is a novel self-supervised task-specific approach to pre-train neural networks for leaf instance segmentation. The results of our experiments also support our key claims, which are: (i) our task-specific pre-training improves the performance on leaf instance segmentation and reduces the amount of labeling required; (ii) distance information and number of points are more important than the use of a second view as in common contrastive learning; (iii) increasing the embedding size boosts the performance of our pre-training more than the performance of the randomly initialized network; and (iv) our novel automatic domain-specific postprocessing achieves better performance with respect to common state-of-the-art methods.

### A. Experimental Setup

**Datasets.** We recorded 3 566 images using a UAV on a 50 m × 46 m field. We compute the point cloud of the field via bundle adjustment. The results of our experiments are reported on a test set, made of 58 point clouds. For pre-training, we use 2 616 point clouds of plants, extracted via the preprocessing steps described in Sec. III-A.

**Training details and parameters.** In all experiments, we use AdamW [43] with weight decay $10^{-6}$ and initial learning rate $2 \cdot 10^{-3}$ for 100 epochs. We use a batch size of 48, thanks to gradient accumulation. We build the graph using $k = 7$ nearest neighbors and a maximum distance of $\tau = 2$ cm. We fine-tune on labeled data with the configuration of the official code by Jiang et al. [11], from now on called PointGroup, and Marks et al. [44].

**Metrics.** We evaluate the leaf instance segmentation using the mean Average Precision (mAP) [45], which is a common choice for instance segmentation tasks.

TABLE I: Results for the leaf instance segmentation with different pre-training approaches and embedding size $D = 3$.

| Pre-training | # views | mAP [%] |
|---|---|---|
| none | — | 36.8 |
| point-to-point | 2 | 41.2 |
| euclidean | 2 | 41.8 |
| graph | 2 | 42.0 |
| graph | 1 | **44.3** |

TABLE II: Results for the ablations on the graph pre-training on the number fo points used, and on the computation of the distances with the Floyd-Marshall (FW) algorithm versus the distance matrix provided by the k-nearest neighbor (kNN) algorithm.

| # views | # points | mAP [%] | |
|---|---|---|---|
| | | kNN | FW |
| 2 | 7 000 | 36.3 | 42.0 |
| 1 | 7 000 | 36.0 | 40.2 |
| 1 | 10 000 | **39.9** | **44.3** |

### B. Spatially Informed Pre-training

The first experiment compares our different pre-training approaches and shows that a spatially informed pre-training is a better initialization for our target task. We refer to point-to-point in Tab. I as the approach following Eq. (4). It makes no use of spatially information and only enforces that the point has the same embedding in the two augmented views. We can see that this pre-training already boosts the performance compared to the random initialization of the network. When including spatial knowledge the performance is even better, confirming that the spatial informed pre-training is more aligned with the leaf instance segmentation task. Using graph distances obtains the best result, as expected considering the limitations of the Euclidean distances presented in Sec. III-C. Results suggest that, when the graph is representative of the point cloud, the graph distance is a good approximation of the geodesic distance on the surface of the plant. This provides a better initialization for the network that then needs fewer iterations or labels to outperform other approaches.

### C. Best Use of Distances and Views

In the second experiment, we investigate if computing the distances on the graph, even if computationally demanding, improves the performance. We also show that the number of points used in the loss computation is crucial to obtain a good initialization, more than using a second augmented view. We compare the pre-training with the distances computed via the Floyd-Warshall algorithm and the pre-training without running the algorithm, i.e., $\widetilde{\mathbf{D}}$ from Eq. (6). Tab. II shows that using one or two views does not impact the improvement gained from the computation of the distances on the whole graph, we gain approx. 4% of mAP in both cases. Using just one view performs worse if all the other parameters stay the same, but decreases the GPU usage, allowing us to use more points in the loss computation. Increasing the number of points closes the gap between the one-view and the two-views approach, outperforms the latter, and shows that using more points is more beneficial than using a two-views approach. It is important to notice that using two views prevents us to increase the number of points due to the memory usage of the second view and all its embeddings. The final outcome of the experiment is that, using the Floyd-Warshall algorithm is worth its computational cost, and that, considering the memory usage of each point and its embedding, using only one view with more points provides a better network initialization.

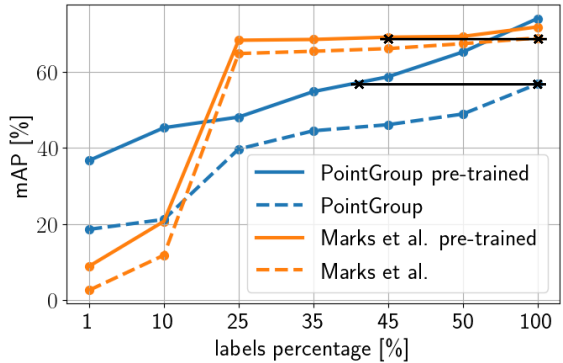

Fig. 4: The plot of the mAP fine-tuning with different amounts of labels for the two approaches [11], [44]. In black we highlight the difference in labels from the best results without pre-training.

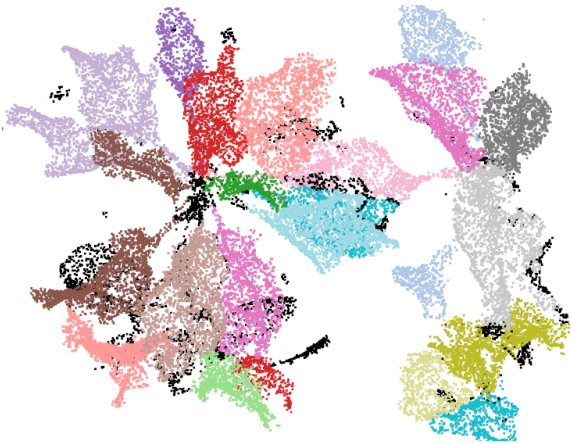

Fig. 5: Results obtained using the work by Marks et al. [44] pre-trained with our method and fine-tuned on 50% of the labeled data.

### D. Label Requirement Reduction

This experiment aims to show the capability of our approach to reduce the required amount of labeled data for the leaf instance segmentation task. We initialize the backbones with our pre-training and use progressively less labels when fine-tuning. The results in Fig. 4 show that we can boost the performance when using all the available data, and we obtain a similar or better mAP using 45% of the labeled data or more. We also include some qualitative images obtained after fine-tuning on 50% of the labeled data in Fig. 5.

### E. Embedding Size Scalability

The following experiments investigate which embedding size is most suitable for the task and if the unsupervised pre-training consistently improves the results. Increasing the embedding size gives us more representational power but also more memory usage for each point.

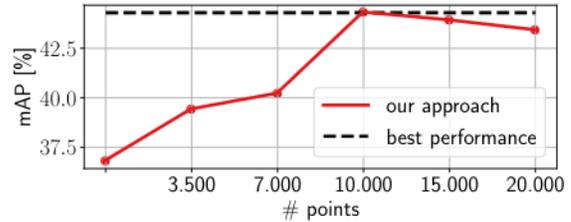

Fig. 6: The plot of the mAP fine-tuning different pre-trainings and number of points used in the loss, with embedding size $D = 3$. The curve shows a linear dependency before saturation occurs.

TABLE III: Results for the mAP on the leaf instance segmentation with different pre-training approaches and embedding sizes D.

| D | # points | mAP [%] | | |
| --- | --- | --- | --- | --- |
| | | without pre-training | euclidean pre-training | graph pre-training |
| 3 | 10 000 | 36.8 | 41.8 | **44.3** |
| 24 | 10 000 | 45.7 | 49.3 | **60.2** |
| 48 | 8 000 | 56.8 | 61.0 | **74.2** |

| Approach | # points | D | mAP [%] |
| --- | --- | --- | --- |
| DepthContrast | 10 000 | 96 | 65.8 |
| SegContrast | 20 000 | 128 | 67.2 |

*1) Number of Points or Embedding Size:* Fig. 6 shows the results of the fine-tuning, after pre-training with the same embedding size and different number of points in the loss computation. We can see that for the first half of the plot, the more points we use, the better the performance is. However, using more than 10 000 points does not further increase the final mAP. This can also be due to the number of available points for each input point cloud $\mathcal{P}_i$, when $|\mathcal{P}_i| < 10\,000$ the loss would use less points, not impacting the final performance.

*2) Increasing the Representational Power:* Tab. III shows the result of the fine-tuning with different embedding sizes for our graph approach, the Euclidean distance-based approach, and the randomly initialized network. We use the same number of points for embedding sizes 3 and 24 to make a fair comparison. The highest embedding size prevented us from using 10 000 points because of the higher memory usage. We are confidently sure that if a more powerful GPU is available, using 10 000 points for the last experiment would improve a bit the performance. However, Tab. III shows the best results we could obtain using the same machine. We can see that using the graph distances is consistently better for all the embedding sizes. Increasing the embedding size yields higher mAP for all the approaches due to the greater representational power. However, we notice that the gap between the graph pre-training and the other approaches is also increasing with the embedding size. This suggests that the graph pre-training is able to learn the plant structure from the points positions, thus providing a better initialization.

### F. Domain-Specific vs. Representational Power

In this experiment, we aim to evaluate our pre-training against state-of-the-art methods, which are not trained on the domain-specific data but have more representational power.

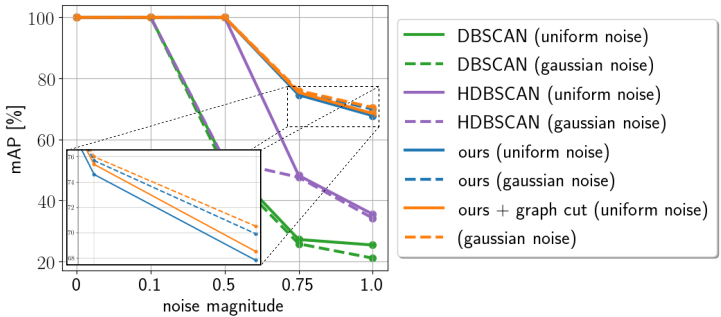

Fig. 7: The mAP [%] of the four postprocessings with increasing noise magnitude (the values stand for the maximum noise allowed) with respect to perfect embeddings.

TABLE IV: Evaluation of the postprocessings with different embedding sizes and pre-training approaches. The numbers in parenthesis represent the difference with respect to HDBSCAN run on the 3D positions only (no network).

| post processing | embedding size | mAP [%] euclidean pt | mAP [%] graph pt |
|---|---|---|---|
| DBSCAN | 24 | 5.7 (+2.9) | 6.9 (+4.1) |
| HDBSCAN | 24 | 6.9 (+4.1) | 11.3 (+8.5) |
| ours | 24 | 7.9 (+5.1) | 12.2 (+9.4) |
| ours + graph cut | 24 | 9.3 (+6.5) | **12.4 (+9.6)** |
| DBSCAN | 48 | 6.3 (+3.5) | 10.9 (+8.1) |
| HDBSCAN | 48 | 7.5 (+4.7) | 11.3 (+8.5) |
| ours | 48 | 8.1 (+5.3) | 12.3 (+9.5) |
| ours + graph cut | 48 | 11.3 (+8.5) | **13.6 (+10.8)** |

DepthContrast [33] uses an embedding size of 96 and 10 000 points for pre-training, while SegContrast [46] uses an embedding size of 128 and 20 000 points. The results in the second part of Tab. III shows that even if the pre-trainings have a higher representational power — twice more for DepthContrast and even more for SegContrast — and use more points — 2 000 and 12 000 respectively — they both fall short if compared to our best result.

### G. Automatic Postprocessing

In the last set of experiments, we evaluate the embeddings we get from our pre-training without fine-tuning. Firstly, we evaluate three different postprocessing algorithms on perfect embeddings (labels) and we progressively add noise, to show how the final mAP degrades. We use DBSCAN [26] and HDBSCAN [47] as baselines. We compare them with our postprocessing, as explained in Sec. III-D, and with a second algorithm implemented by us and based on graph cuts [48]. This variant uses the same Step 1 of our postprocessing. In Step 2 we use the graph cut operation to separate one leaf from the rest of the plant. This must be repeated for each cluster, i.e., leaf, found in the first step. Step 3 merges the clusters into the final prediction.

After assessing the performance of the different postprocessings, we evaluate the fully unsupervised approach.

*1) Postprocessings Evaluation:* Fig. 7 shows the results for the four postprocessings using perfect embeddings and adding noise. We conduct two experiments, adding (1) random noise over all the samples or (2) gaussian-shaped noise with a higher magnitude near the center of the plants, according to what we discussed in Sec. III-D. Both of our postprocessings outperform the baselines (approximately $+50\%$ of mAP over DBSCAN and $+35\%$ over HDBSCAN), especially when adding Gaussian-shaped noise. We can see that using graph cut performs slightly better (approximately $+0.6\%$ of mAP), but the need to build the graph does not scale well for high-resolution point clouds. The results suggest that our postprocessings are more robust to the expected noise than usual clustering algorithms.

*2) Fully Unsupervised Embeddings Evaluation:* In Tab. IV we report the results for the postprocessing algorithms on the embeddings predicted from our unsupervised approaches, both Euclidean and graph-based. In the parenthesis, we provide the difference with respect to the results obtained from HDBSCAN on the points position, which we consider the most basic geometric baseline. Our best performance is less than 10% worse than DBSCAN on the noisy but perfect embeddings. The results confirm that the graph approach extracts more meaningful features and that our postprocessing has better performance than state-of-the-art approaches. This set the mAP for the task, without the use for labels, from 2.8% of mAP to 13.6%.

## V. CONCLUSION

In this paper, we presented a novel approach to pre-train a neural network in a domain- and task-specific self-supervised fashion for leaf instance segmentation, and we proposed an automatic embedding-based postprocessing. Our method exploits the large amount of data that is easy to collect and tries to reduce the labeling effort required to obtain state-of-the-art performance on the leaf instance segmentation task. The approach relies on domain-specific data augmentations and a task-specific loss, plus domain-specific automatic postprocessing. We implemented and evaluated our approach, provided comparisons to other pre-training approaches, and supported all claims made in this paper. The experiments suggest that our graph pre-training is a better initialization for the target task and it boosts the final performance over all the scenarios. We achieve better performance when using the same amount of data and computational power, and we can achieve the same performance using fewer resources. The results show that our automatic domain-specific postprocessing outperforms common state-of-the-art clustering algorithms, being also more robust to both uniform noise and Gaussian-shaped noise expected from the task. Our approach will enable robotic systems to perform crop monitoring in a more efficient way, reducing the requirements for labels and improving the task performance.